\documentclass[11pt, letterpaper]{article}

\usepackage[final]{microtype}
\usepackage{color}
\usepackage[ruled]{algorithm2e}
\usepackage{subcaption}
\usepackage{varwidth}
\usepackage{capt-of}
\usepackage[blocks]{authblk}
\usepackage{boldline}
\usepackage{epsfig}
\usepackage{latexsym}
\usepackage{url}
\usepackage{float}
\usepackage[hypertexnames=false,hyperfootnotes=false]{hyperref}
\usepackage{texnansi}
\usepackage{tikz}
\usepackage{afterpage}
\usepackage{enumerate}
\usepackage[normalem]{ulem}
\usepackage{lmodern}
\usepackage{rotating}
\usepackage[ruled]{algorithm2e}
\usepackage[noend]{algpseudocode}
\usepackage{graphicx}
\usepackage{caption}
\usepackage{subcaption}
\usepackage{booktabs}

\usepackage{wrapfig} 
 
\usepackage{sectsty}

\usepackage{ifthen}

\usepackage[leqno]{amsmath}
\usepackage{amsthm}
\usepackage{amssymb}
\usepackage{amsfonts}
\usepackage[margin=10pt,font=small,labelfont=bf]{caption}

\usepackage{enumitem}
\setlist[enumerate]{topsep=1pt,itemsep=1pt,partopsep=1pt,parsep=1pt}

\usepackage{dsfont}

\usepackage{natbib}
 \bibpunct[, ]{(}{)}{,}{a}{}{,}%

\makeatletter
\g@addto@macro{\UrlBreaks}{\UrlOrds}
\makeatother

\newcommand{\field}[1]{\ensuremath{\mathbb{#1}}}
\newcommand{\N}{\ensuremath{\field{N}}} 
\newcommand{\R}{\ensuremath{\field{R}}} 
\renewcommand{\S}{\ensuremath{\field{S}}} 
\newcommand{\PR}{\ensuremath{\mathsf{P}}} 
\newcommand{\E}{\ensuremath{\mathsf{E}}} 

\newcommand{\Hscr}{\ensuremath{\mathcal H}}

\newcommand{\Mscr}{\ensuremath{\mathcal M}}

\newcommand{\Vscr}{\ensuremath{\mathcal V}}

\DeclareMathOperator*{\argmax}{\mathrm{argmax}}

\newtheoremstyle{thm-sf}{}{}{\itshape}{}{\sffamily\bfseries}{.}{ }{}
\theoremstyle{thm-sf}

\newtheorem{assumption}{Assumption}
\newtheorem{definition}{Definition}

\newtheorem{theorem}{Theorem}

\newtheorem{lemma}{Lemma}

\newtheorem{proposition}{Proposition}

\renewcommand{\proofname}[1]{{\normalfont\sffamily\bfseries #1}}

\newenvironment{myproof}[1][\proofname]{%
  \proof[\normalfont\sffamily\bfseries #1]%
}{\endproof}
\newcommand\numberthis{\addtocounter{equation}{1}\tag{\theequation}}

\newcommand{\OPT}{\mathrm{OPT}}
\newcommand{\ANN}{\mathrm{ANN}}
\newcommand{\LSS}{\mathrm{LSS}}

\newcommand{\Sf}{\ensuremath{\field{S}}}

\usetikzlibrary{arrows,patterns,plotmarks}
\tikzstyle{every picture} += [>=stealth]
\allsectionsfont{\sffamily}

\makeatletter
\def\@seccntformat#1{\csname the#1\endcsname.\quad}
\makeatother

\floatstyle{ruled}
\provideboolean{fastcompile}
\newcommand{\hidefastcompile}[1]{\ifthenelse{\boolean{fastcompile}}{}{#1}}

\usepackage[margin=1in]{geometry}
\usepackage[nodisplayskipstretch]{setspace}
\newtheorem*{le:ell}{Lemma~\ref{le:ell}}
\newtheorem*{le:feas}{Lemma~\ref{le:feas}}
\newtheorem*{le:sample_complexity}{Lemma~\ref{le:sample_complexity}}

\usepackage{setspace}

\title{\textsf{\textbf{Optimizing Offer Sets in Sub-Linear Time}}}
\date{}

\begin{document}

	\author[1]{Vivek F. Farias}
	\author[2]{Andrew A. Li}
	\author[3]{Deeksha Sinha}
	\affil[1]{Sloan School of Management, Massachusetts Institute of Technology}
	\affil[2]{Tepper School of Business, Carnegie Mellon University}
	\affil[3]{Operations Research Center, Massachusetts Institute of Technology}

\maketitle
\setstretch{1.55}

\begin{abstract}
Personalization and recommendations are now accepted as core competencies in just about every online setting, ranging from media platforms to e-commerce to social networks. While the challenge of estimating user preferences has garnered significant attention, the operational problem of using such preferences to construct personalized offer sets to users is still a challenge, particularly in modern settings where a massive number of items and a millisecond response time requirement mean that even enumerating all of the items is impossible. Faced  with such settings, existing techniques are either (a) entirely heuristic with no principled justification, or (b) theoretically sound, but simply too slow to work.

Thus motivated, we propose an algorithm for personalized offer set optimization that runs in time {\em sub-linear} in the number of items while enjoying a uniform performance guarantee. Our algorithm works for an extremely general class of problems and models of user choice that includes the mixed multinomial logit model as a special case. We achieve a sub-linear runtime by leveraging the dimensionality reduction from learning an accurate latent factor model, along with existing sub-linear time approximate near neighbor algorithms. Our algorithm can be entirely data-driven, relying on samples of the user, where a `sample' refers to the user interaction data typically collected by firms. We evaluate our approach on a massive content discovery dataset from Outbrain that includes millions of advertisements. Results show that our implementation indeed runs fast and with increased performance relative to existing fast heuristics.

\end{abstract}

\section{Introduction} \label{secIntro}

A common problem in modern web-services revolves around constructing personalized offer sets (or assortments) for users with the goal of optimizing some objective function related to each specific user's experience with the service. Recommendation problems (such as those faced by services like Netflix) represent a canonical version of such a problem. Assortment optimization problems (as faced by online retailers) are another common example. Now, in practice, rigorous latency constraints are an important consideration when building algorithms to construct optimized offer sets. For example, a large-scale study (\cite{akamai2017online}) recently showed that a 100 millisecond delay in loading page content can result in a decrease in conversion (i.e. consumption, purchase, etc.) of up to 7\%. When taken together with the vast size of the product universe (which can run to the tens of millions), such constraints place severe limitations on what a {\em real-time} algorithm for constructing an optimized offer can do in practice. As such, practically implementable algorithms for real-time offer set optimization at scale should ideally exhibit runtimes {\em sub-linear} in the size of the universe of potential products. 

The dynamic nature of the product universe limits the use of pre-computation. Moreover, as will be evident later, distributional models of the user further restrict our ability to exhaustively pre-compute optimal offer sets. As a result, the dominant approach today to building optimized offer sets in real-time relies on the design of so-called approximate nearest neighbor (ANN) algorithms. Succinctly, such algorithms leverage metric-space representations of users and products wherein the distance between a user and product is inversely related to how attractive the product is to the user. Whereas such a metric space representation of products may be constructed offline, the real-time problem then consists of identifying a point in the metric space that corresponds to the user, and finding the $k$ products in the metric space that are nearest to that user in time sub-linear in the number of products. This problem is well-solved both theoretically and practically. 

Now an economically grounded approach to constructing an optimal offer-set would typically rely on modeling the user's utility for various products. An offer set constructed assuming that a user made choices to maximize utility would then seek not just to pick products that are likely interesting to the user, but would further seek to account for substitution and complementarity effects. There is by now a vast literature dedicated to the estimation of models of choice as well as the associated assortment optimization problems given such models. The assortment optimization algorithms developed in this context are, while typically efficient, not sub-linear. On the other hand, the ANN paradigm while sub-linear is typically unable to account for assortment effects in a principled fashion, which typically results in a slew of ad-hoc algorithmic tweaks. Here, we seek to begin bridging this gap. 

\subsection{Our Contributions} The present paper seeks to develop sub-linear time algorithms for offer-set optimization while allowing for rich, economically grounded models of customer choice. Specifically, like is typical in the ANN paradigm, we are endowed with a metric space. We are given a universe $V$ of $n$ products, wherein each product $v \in V$ is represented as a fixed point in the metric space (we will describe later on common ways for estimating such an embedding). Similarly, a user $U$ is a {\em random} point in this space; allowing for $U$ to be random is key to modeling user behavior in a way that is congruent with established models of choice.

Our objective then is to solve, in sub-linear time, a problem of the form
\[
\max_{S \subset V, |S| \leq k} 
\E \left[
f(S, U)
\right], 
\]  
where the decision space is the subsets of products of cardinality at most $k$, and the expectation is over $U$. The principle assumptions we place on the functions $f(\cdot, U)$ are that we require these functions be sub-modular, and further that $f(\{v\}, U)$ be non-increasing in the distance $d(U,v)$. As we discuss later, this framework is quite flexible: for instance, it immediately captures the problem of picking an offer set to maximize conversion (i.e. the probability of a purchase) where consumer choice is driven by an essentially arbitrary random utility model. 

With respect to this model, we make the following contributions:
\begin{enumerate}
\item {\bfseries {\sffamily A Sub-linear Time Algorithm:}} Our primary contribution is a sub-linear time algorithm to solve the optimization problem above with uniform performance guarantees. Our algorithm relies on a procedure for constructing, in sub-linear time, a particular sub-linear sized subset of products. This set enjoys the property that the optimal value of our optimization problem restricted to this set is close to the optimal value of the optimization problem over all products. As such, we then simply solve our optimization problem over this restricted set. A greedy algorithm trivially guarantees both a sub-linear run-time and a constant factor approximation. 

\item {\bfseries {\sffamily A New Sampling Scheme:}} Our key algorithmic contribution is our approach to constructing the sub-linear set of candidate products, which we dub {\em locality-sensitive sampling}. Locality-sensitive sampling is a simple idea motivated by the same locality-sensitive hash functions that underly ANN algorithms. By re-interpreting the standard near neighbor problem as one of {\em sampling} items according to a specific decreasing function of their distances from a query point, we are able to solve the same problem for arbitrary decreasing functions. This generalized sampling problem, along with our sub-linear time solution, may be of independent interest.
	
\item {\bfseries {\sffamily Empirical Evaluation:}} We present an empirical study on a large-scale corpus of real page-view data from the online advertising platform Outbrain. The dataset contains two billion page views of seven hundred million unique users. Our experiments establish the value of our procedure over existing sub-linear time heuristics, and in particular, that (a) our model of user choice is more accurate in predicting user behavior than the models implicitly assumed by these heuristics, and (b) our algorithm outperforms these heuristics in terms of conversion rate. 
	
\end{enumerate}

The rest of this paper is organized as follows: we review related work in the remainder of this section. Section 2 introduces our problem formally, along with our key modeling assumptions. Section 3 describes the motivation for our algorithm by way of an idealized sampling procedure. We then describe our actual algorithm, which is designed to approximate this idealized procedure, in Section 4. Experimental results are described in Section 5, and finally conclusions are drawn in Section 6.

\subsection{Related Work}

This work is related to three existing streams of literature, as we describe now. 
\paragraph{Recommendation Algorithms:} In the area of recommender systems, the problem of learning user preferences from previous interactions has been studied extensively (\cite{jin2003collaborative,jin2002preference,freund2003efficient,schapire1998learning}). For the most part, successful learning algorithms work by embedding both users and items within some metric space such that a user's affinity toward an item is inversely related to their pairwise distance.
See \cite{adomavicius2005toward} for an extensive survey of content-based, collaborative and hybrid recommendation approaches, and \cite{zhang2019deep} for a survey of modern approaches based on deep learning.

A recent problem in this stream of literature is how to capture the impact of {\em diversity} in recommendations. These efforts have mostly focused on quantifying and maximizing diversity in recommendations sets. \cite{kunaver2017diversity} is an extensive survey of the research in this area. A key limitation of the current research here is that the diversity metric is not standardized. Moreover, increasing diversity has often been viewed as sacrificing accuracy of the recommendation set. We will take a more systematic approach to this. 

\paragraph{Assortment Optimization:}
Another stream of literature related to our work is assortment optimization in the field of operations management. Assortment optimization is a principled modeling approach to choosing an optimal assortment to offer to customers. 
\cite{kok2008assortment} provides an  overview of  models found in literature and  approaches common in practice. Integral to the assortment optimization problem is the model for user choice.  One of the most well studied  and  commonly used choice model is the Multinomial Logit (MNL) model.  The  assortment optimization problem with the MNL choice model is tractable, even  under various constraints (\cite{talluri2004revenue, rusmevichientong2010dynamic, davis2013assortment}). 
Though being attractive due to its tractability, the MNL models suffers from  Independence of Irrelevant Alternatives (IIA) property.  
To overcome the IIA limitation,  the Nested Logit  (\cite{williams1977formation}) and Mixed Multinomial Logit models were proposed. More recently, assortment optimization has been studied under some new choice models like  the Markov chain choice model (\cite{desir2015capacity}),  distance-comparison based choice model (\cite{kleinberg2017comparison}), distribution over rankings (\cite{farias2013nonparametric}) and its variations (\cite{desir2016assortment}). One limitation of this stream of work is that sub-linear time algorithms effectively do not exist. Even linear time algorithms are rare and restricted to models like the simple multinomial logit that fail to capture user diversity.

\paragraph{Approximate Nearest Neighbors:}
The third stream of literature that is relevant to our work is the problem of nearest neighbor (NN) search.  In this problem, the goal is to pre-process the given data set so that the  nearest neighbor to a query can be efficiently calculated.
 \cite{chavez2001searching} give an overview of methods that have been  proposed to solve this problem. Some sample works on the NN search problem are \cite{omohundro1989five,  sproull1991refinements,  bentley1975multidimensional}, and \cite{yianilos1993data}.  %
 We focus on a particular type of approximate nearest neighbor search algorithm called Locality Sensitive Hashing (LSH) (\cite{andoni2008near}).   \cite{pauleve2010locality} describe various hash functions used in LSH algorithms. 
 These have been used in several applications, 
 but in particular find themselves used extensively in recommendation systems. This application has largely focused on obtaining binary representations  of users and items which can then be used for doing fast similarity search computations (\cite{karatzoglou2010collaborative, zhou2012learning, liu2014collaborative, das2007google, liu2018discrete, zhang2014preference}).

\section{Model and Assumptions}
We begin by introducing the core optimization problem that will be the subject of the rest of this paper. The problem is to select a personalized offer set that maximizes expected reward, subject to a cardinality constraint. Let $V$ denote the universe of items or products we can offer, and $k \in \N$ the maximum cardinality allowed, meaning the set of feasible offer sets is $\{S \subset V: |S| \le k\}$. The need for personalization is driven by the notion of a user `type': we assume that each user has a type, which takes values in some set $\Mscr$, and that this type governs the reward we obtain for offering a given offer set. That is, the reward function, which we denote $f(\cdot,\cdot)$, is a map from $2^V \times \Mscr$  to $[0,1]$, where w.l.o.g. the reward is bounded above by 1. To fix a concrete running example, consider the problem of online content recommendation: the items are webpages, and $f(S,u)$ is the {\em conversion} probability, i.e. the probability that a user of type $u$ will visit at least one of the webpages in $S$ if they are recommended together -- we will expand on this example later in this section.

To summarize so far, if we were given a user of type $u \in \Mscr$, we would seek to solve the problem $\max_{S \subset V, |S| \le k} f(S,u)$. We will see later that this problem is `easy' in many reasonable settings. Instead, one of the two primary challenges we seek to address in this paper is how to deal with {\em user heterogeneity}, i.e. when the user type is not known exactly ex-ante. We will assume that this uncertainty is modeled as a random variable $U$ over $\Mscr$, whose distribution we know. Our goal then is to solve the following stochastic optimization problem:
\begin{equation} \label{eqnOPT}
	\OPT \equiv \max_{S \subset V, |S| \le k} \E \left[ f(S,U) \right].
\end{equation}
For the rest of this paper, we will take $U$ to be uniformly distributed over $m$ types: $u_1,\ldots,u_m \in \Mscr$. There are two motivations for this: first, for $m$ sufficiently large, this assumption is  without loss, as replacing the expectation in \eqref{eqnOPT} with a sample average approximation results in negligible loss. Lemma \ref{lemSAA} in Appendix \ref{secAppendixSAA} shows that $m = \Omega(k\log n)$ is sufficiently large, where $n \equiv |V|$ is the number of items. Second, in practice, what is quite often done to model $U$ is that past observations of a given user are mapped to points in $\Mscr$, and $U$ is taken to be a distribution whose support is over these points; the uniform distribution is one natural choice (we will provide experimental evidence for this in Section \ref{secExperimentModel}).

As was described in the Introduction, we seek to solve \eqref{eqnOPT} in online settings in which the number of items $n$ is massive and the optimization must be performed extremely fast, often so fast that even algorithms linear in $n$ are too slow. Thus, the second primary challenge we face is to solve, or approximate, problem \eqref{eqnOPT} in {\em sub-linear} time: $o(n)$. This will require three assumptions, which we will describe in detail in the remainder of this section. Precisely, imposing these assumptions will allow us to guarantee (in expectation) an approximation of $\OPT$ using a randomized algorithm whose expected runtime, amortized over multiple users, is $O(n^{1-\epsilon})$ for some strictly positive $\epsilon$.

Our first assumption is that the reward function for any user type be monotone submodular:
\begin{assumption}\label{assuSubmodular}
For every $u \in \Mscr$, the function $f(\cdot,u)$ is monotone submodular.	
\end{assumption}
The set of reward functions satisfying Assumption \ref{assuSubmodular} is rich enough to include the conversion function for recommendation problems and a subclass of assortment optimization problems against the mixed multinomial logit choice model. Making this assumption is the first step in achieving sub-linearity. In fact, Assumption \ref{assuSubmodular} already implies that $(1-1/e)\OPT$ can be guaranteed in {\em linear}  time, as the greedy algorithm is $(1-1/e)$-optimal for maximizing monotone submodular functions subject to cardinality constraint (\cite{nemhauser1978analysis}). Since sums of monotone submodular functions are monotone submodular, the greedy algorithm for \eqref{eqnOPT}  achieves $(1-1/e)\OPT$. Our eventual algorithm will achieve a strictly lower (but still constant) approximation guarantee, but will improve on the greedy algorithm's $O(kmn)$ runtime.

\subsection{User and Item Embedding}
The remaining two assumptions we make will allow us to use the machinery of approximate nearest neighbor algorithms in order to improve from linear to sub-linear time. To discuss these, we will first need to describe the underlying geometry of our problem. Recall that we model user types as elements of a set $\Mscr$, which so far is an arbitrary set. We will assume that $\Mscr$ is in fact a metric space, equipped with a metric denoted by $d(\cdot,\cdot)$. We will also assume that our universe of items is embedded in this same space: $V = \{v_1,\ldots,v_n\} \subset \Mscr$. 

Such embeddings are ubiquitous in predictive algorithms for personalization which, by and large, operate by estimating feature (or latent-factor) representations of users and items so that, loosely speaking, a user will have a stronger preference for items whose features `align' more closely to his or her own features (or equivalently, those items whose features are closer in distance with respect to a carefully calibrated metric). The metric space will often either be Euclidean space or the unit-ball in Euclidean space, but using the Euclidean metric is not a requirement. Instead, what we will need to assume about our space is that there exists an appropriate data structure that returns approximate near neighbors in sub-linear time:
\begin{assumption}\label{assuANN}
For any distance $\gamma > 0$, and constants $c > 1$, $\beta \in [0,1)$, and $\epsilon \in (0,1]$, there exists a (randomized) data structure $\ANN[V,\gamma,c,\beta,\epsilon]:\Mscr \to 2^V$, and a corresponding $\alpha \equiv \alpha(c,\beta,\epsilon) < 1$ such that, given any query point $u \in \Mscr$: 
\begin{enumerate}
\item If \[ \sum_{v \in V}\mathds{1}(d(v,u) \le c\gamma) \le n^\beta, \]
then for each $v \in V$  such that $d(v,u) \le \gamma$, \[\PR\left(v \in \ANN[V,\gamma,c,\beta,\epsilon](u)\right) \ge 1-\epsilon.\]
\item The runtime of querying this data structure is $O(n^\alpha)$.
\end{enumerate}
Here, constants suppressed by the big-Oh notation depend only on $\Mscr$ (e.g. dimensionality).
\end{assumption}
Assumption \ref{assuANN}, while perhaps unusual at first, is stated in the form typically taken in theoretical guarantees for approximate near neighbor algorithms. In words, the data structure assumed here takes any point in $\Mscr$ as input, and outputs every item in $V$ that is within a pre-specified distance $\gamma$ of the input, each with sufficiently high probability. Most importantly, the runtime of this operation is sub-linear (since $\alpha$ is assumed to be less than 1), assuming that the number of these near neighbors itself is sub-linear (since $\beta$ is assumed to be less than 1). As a sanity check, if $\epsilon$ could be taken to be $0$, and $c$ could be taken to be $1$, this would correspond to an exact near neighbor algorithm. Having $\epsilon > 0$ reflects the fact that near neighbors are only guaranteed to be returned with high probability. Having $c > 1$ reflects that in the process, elements of $V$ slightly further than $\gamma$ are generated as candidates and need to be pruned. 

The study of approximate near neighbor algorithms has produced data structures satisfying Assumption \ref{assuANN} for a variety of metric spaces, including Euclidean space. As we will review later on, one way of construct such a structure uses a family of so-called locality-sensitive hash functions with certain `nice' properties. Our ability to solve \eqref{eqnOPT} in sub-linear time will rely on our ability to sample from a certain distribution on $\Mscr$. The algorithm we develop will rely on a carefully constructed ensemble of data structures of the type defined by Assumption~\ref{assuANN}

Finally, while Assumptions \ref{assuSubmodular} and \ref{assuANN} deal with the reward function $f$ and the underlying metric space $(\Mscr,d)$ separately, there has so far been nothing rigorously tying the two together which would allow us to leverage the metric structure. This is the purpose of our final assumption, which states that the distance between the embeddings of a user and an item directly encodes the corresponding reward for offering that item alone to that user:
\begin{assumption} \label{assuSublinear} 
There exists a non-increasing function $p:\R^+ \to [0,1]$ such that
\[   p(d(v,u)) \ge f(\{v\},u) \;\; \text{for each } \; u \in \Mscr,\; v \in V. \]
In addition, there exists some $\beta \in [0,1)$ and $c > 1$ such that
\[ \sum_{v \in V} p\left(\frac{d(v,u)}{c}\right) \le n^\beta  \;\; \text{for each } \; u \in \Mscr. \]
\end{assumption}
The function $p(\cdot)$ captures the inverse relation between distance in $\Mscr$ and reward, and will play a crucial role in our algorithm. In particular, we will treat $p(\cdot)$ as a probability in a sampling-based approach. 
The second part of Assumption \ref{assuSublinear}, which will allow us to make use of the approximate near neighbor in sub-linear time, assumes that for all user types $u$, the total reward gotten by offering each item individually is sub-linear. 
This may, for example, reflect the fact that users' appetites for content are not limited by a lack of items, but rather a limit in time, attention, etc. Additionally, this condition is required to be robust in the following sense: there exists some $c > 1$ such that the condition above still holds if each $v \in V$ is replaced by a contracted vector $\tilde{v}$ such that $d(\tilde{v}, u) = d(v, u)/c$.

\subsection{Examples}
We conclude this section by describing two common models which fit the framework we have outlined and satisfy our three assumptions.

\paragraph{Conversion Under Random Utility Choice Models:} As described previously, the goal in recommendation problems is typically to induce {\em conversion}, i.e. selecting at least one of the items in the offer set (e.g. clicking on one of a set of web links, or listening to one of a list of songs). In this setting, $f(S,u)$ is a {\em conversion function} which, for any set of items $S \subset \Vscr$, is the probability of conversion when customer $u$ is offered set $S$. 

Random utility models are commonly used to describe user choice behavior. One generic way of employing these models in the conversion problem is to assume that $f(S,u)$ takes the form 
\begin{equation} \label{eqnFS} 
f(S,u) = \PR \left(\max_{v \in S} ( \mu(d(v,u)) + \epsilon_v) > \epsilon_{\emptyset}  \right),	
\end{equation}
where $\mu:\R^+ \to \R^+$ is a non-increasing function, and the $\epsilon_v$'s and $\epsilon_{\emptyset}$ are i.i.d. mean-zero random variables. Here, $\mu(d(v,u)) + \epsilon_v$ is the random utility associated with selecting item $v$, with $\mu(\cdot)$ translating distance to a mean utility, and $\epsilon_v$ capturing idiosyncratic noise. The random utility of selecting no item is $\epsilon_{\emptyset}$, assuming w.l.o.g. that the mean utility of this option is zero. 
Users then choose the option (either one of the recommended items, or no item) that maximizes their utility, and \eqref{eqnFS} is the probability that the utility of any recommended item is higher than the utility of selecting nothing. 

The formulation in \eqref{eqnFS} satisfies Assumption 1 immediately, and the function $p(\cdot)$ required by Assumption 3 can be constructed from the distribution of the $\epsilon$'s. This setup is extremely general and encodes many popular choice models.
For example, taking the $\epsilon$'s to be Gumbel random variables yields the multinomial logit model, and allowing for random $U$ yields the mixed multinomial logit.

One commonly used choice of metric space and utility function in the conversion problem is the following: 
let $\Mscr = \S^{d-1}$, i.e. the unit ball in $d$-dimensional Euclidean space. As stated earlier, this space satisfies Assumption 2. Now since we are dealing with unit-vectors, we have that $d(v,u)^2 = 2(1-v^\top u)$, so taking $\mu(x) = 1-x^2/2$ yields a natural form for the mean utility:
\[ \mu(d(v,u)) = v^\top u. \]
Finally, it is worth noting that this formulation is compatible with a number of approaches to constructing metric space representations of users and products,  ranging from simple logistic regression, to collaborative filtering, to state of the art approaches such as factorization machines (\cite{rendle2010factorization}) and field-aware factorization machines (\cite{juan2016field}), the latter having been a key component in the winning entries of three major recent public prediction competitions.

\paragraph{Assortment Optimization Under the Mixed Multinomial Logit Model:}
In operations management, a classic problem is to select an assortment of products to offer to customers so as to maximize expected {\em revenue}. Changing the objective from conversion to revenue yields a far more difficult problem, especially given the rich set of choice models and additional operational constraints one could assume. Our model and algorithm will in no way offer a completely general sub-linear time solution, but there do exist meaningful instances to which they can be applied, as we illustrate by example now.

Let $r_j$ be the revenue gained if a customer purchases product $v_j$. Here, we will just work out the setting where the underlying choice model is the multinomial logit:
\begin{equation} \label{eqnAO}
f(S,u) = \sum_{v_j \in S} r_j \frac{\exp(v_j^\top u)}{w+\sum_{v \in S}\exp(v^\top u)},	
\end{equation}
where $w \ge 0$ is a parameter that controls the likelihood that no product is selected.
The objective in \eqref{eqnAO} is not in general monotone or submodular, but there are a variety of conditions which imply both. For example, one such condition shown in \cite{han2019assortment} is if the minimum and maximum revenues (denoted $r_{\min}$ and $r_{\max}$) are not too far apart: 
\[ \frac{r_{\min}}{r_{\max}} \ge \max_{S \subset V, |S| \le k} \sum_{v_j \in S} \frac{\exp(v_j^\top u)}{w+\sum_{v \in S}\exp(v^\top u)}. \]
The expression on the right-hand side is equal to the maximum conversion (as defined previously) probability for a user of type $u$ across all feasible assortments. In particular, the revenues are allowed to vary more when this quantity is small, or equivalently, when $w$ is large. The required upper bound on $f(\{v\},u)$ can be gotten by treating all revenues as $r_{\max}$.

\section{Algorithm Overview}
Before describing our approach, we could first consider whether some sort of brute force pre-computation would suffice, that is, simply solving \eqref{eqnOPT} in advance for a sufficiently comprehensive set of distributions $U$. If feasible, this would certainly qualify as an amortized sub-linear (constant, in fact) time algorithm. There are at least two reasons why this approach might be infeasible or at best impractical. First, the size of a `comprehensive' set of distributions $U$ could be massive -- even having made our reduction so that $U$ is uniformly distributed over $m$ points of $V$, this set is of size $O(n^m)$ -- in which case it may be practically impossible to compute and/or store it. Second, in almost all settings, the product set is dynamic. For example, the universe of online content is constantly changing. Thus, the data structure needs to be dynamic, ideally capable of fast additions and deletions. The structure we describe in the next section will allow these dynamic updates in sub-linear time; brute force pre-computation would not.

At a high level, our algorithm proceeds in two steps: 
\begin{itemize}
\item[(a)] Randomly sample a sub-linear sized subset of $V$, which we will denote by $\tilde{V}$, such that if \eqref{eqnOPT} is solved over $\tilde{V}$ instead of $V$, we are still guaranteed a constant fraction $(1-\epsilon)$ of $\OPT$ in expectation. 
\item[(b)] Approximately solve \eqref{eqnOPT} over $\tilde{V}$ using the greedy algorithm. 
\end{itemize}
The crux of our algorithm is the ability to perform step (a) in sub-linear time. 
Assuming that step (a) could be performed in sub-linear time, the greedy algorithm in step (b) would then also run in sub-linear time, and the algorithm as a whole would be guaranteed $(1-1/e)(1-\epsilon)$ of $\OPT$ in expectation. 

Ignoring the runtime of step (a) for a moment, we will first describe an idealized random sampling scheme over the items of $V$ that would return a random subset $\tilde{V}$ that is both sub-linear in size and guaranteed (in expectation) to preserve a constant fraction of $\OPT$ (less a small additive error) when optimized over. To ease notation, we will fix a distribution $U$ and denote the objective function of \eqref{eqnOPT} by $g(S) = \E[f(S,U)]$. 

Suppose that we could randomly sample $\tilde{V}$ such that 
\begin{equation} \label{eqnIdeal}
\PR(v \in \tilde{V}) = g(\{v\}) \;\; \text{for each }  v \in V.
\end{equation}
That is, the likelihood of any item being included in $\tilde{V}$ is equal to the reward that the item would yield when offered alone (recall that this reward is assumed w.l.o.g. to lie in $[0,1]$). The following Lemma shows that making sufficiently many independent draws from such a sampling distribution, and taking the union of these draws, would result in a subset of $V$ that is guaranteed a constant fraction of $\OPT$ if subsequently optimized over:

\begin{lemma} \label{lemPrune}
For $c \in (0,1]$, let $\tilde{V}$ be a random variable taking values in $2^V$ such that
\[ \PR(v \in \tilde{V}) \ge c g(\{v\}) \;\;\; \text{for each }  v \in V,  \]
and for $s \in \mathbb{N}$, let $\tilde{V}_s$ denote the union of $s$ sets drawn i.i.d. from this distribution.

Let $S^*(\tilde{V}_s)$ be an optimal solution to:
\[\max_{S \subset \tilde{V}_s, |S| \le k} g(S).\]
Then for any $\epsilon_1,\epsilon_2 \in (0,1]$, if
\[s \ge \frac{k}{c\epsilon_2}\log\frac{k}{\epsilon_1},\]
then we have
\[ \E[g(S^*(\tilde{V}_s))] \ge (1-\epsilon_1)\OPT - \epsilon_2. \]
\end{lemma}

\begin{myproof}[Proof of Lemma \ref{lemPrune}]

Fix any $\delta \in [0,1]$ (we will tune this quantity in the end).	For any $v \in V$ such that $g(\{v\}) \ge \delta$, we have
	\begin{align*}
\PR(v \notin \tilde{V}_s) 
& = \PR(v \notin \tilde{V})^s \\
& \le \left(1 - c g(\{v\})\right)^s \\
& \le \left(1 - c\delta \right)^s \\
& \le e^{-sc\delta}, \numberthis \label{eqnProof1}
	\end{align*}
where the first equality follows from the definition of $\tilde{V}_s$, and the first two inequalities are by assumption.

	Now we fix any optimal solution $S^*$ to the full problem \eqref{eqnOPT}, and divide it into two disjoint sets: 
	\[S_1 = \{ v \in S^* : g(\{v\}) \ge \delta  \}   \;\; \text{and} \;\; S_2 = \{ v \in S^* : g(\{v\}) < \delta  \}.   \]
Then we have	
	\begin{align*}
	g(S^*(\tilde{V}_s))  
		& \ge g(S_1 \cap \tilde{V}_s) \\ 
		& \ge \PR(v \in \tilde{V}_s \;\; \forall \; v \in S_1) g(S_1) \\
		& \ge \left( 1 - \sum_{v \in S_1} \PR( v \notin \tilde{V}_s) \right) g(S_1) \\
		& \ge \left( 1 - ke^{-sc\delta} \right) g(S_1) \\
		& \ge \left( 1 - ke^{-sc\delta} \right)\left(g(S^*) - g(S_2) \right) \\
		& = \OPT - ke^{-sc\delta}\OPT - \left( 1 - ke^{-sc\delta} \right)g(S_2), \numberthis \label{eqnProof2} 
	\end{align*}
	where the first line is due to the optimality of $S^*(\tilde{V}_s)$ among all solutions contained in $\tilde{V}_s$, the third line is a union bound, the fourth line is due to \eqref{eqnProof1}, and the fifth line is due to submodularity.

To conclude, it will suffice to upper bound the second and third terms in \eqref{eqnProof2} by $\epsilon_1\OPT$ and $\epsilon_2$, respectively. To do this, we choose $\delta = \epsilon_2/k$. For the second term, applying this choice of $\delta$, along with our condition on $s$, yields the following bound:
\[ ke^{-sc\delta}\OPT \le ke^{-\log(k/\epsilon_1)}\OPT = \epsilon_1 \OPT. \]
For the third term, by submodularity and the definition of $S_2$,  
\[ \left( 1 - ke^{-sc\delta} \right)g(S_2) \le g(S_2) \le \sum_{v \in S_2}g({v}) \le k\delta = \epsilon_2. \]

\end{myproof}

Lemma \ref{lemPrune} shows that to approximate $\OPT$ to arbitrary precision in expectation, it suffices to sample $s = O(k \log k)$ times from a distribution approximately satisfying \eqref{eqnIdeal}. In fact, the Lemma states that the sampling probabilities do not need to match \eqref{eqnIdeal}, but that they just need to be lower bounded by some constant fraction $c$ of \eqref{eqnIdeal}. In the algorithm we outline later, we will arbitrarily take this fraction to be $c=1/2$.

Having guaranteed that a constant fraction of $\OPT$ is preserved, the other required condition on \eqref{eqnIdeal} is that the resulting subset be sub-linear in size, as the greedy algorithm that follows is linear in the size of this set. Fortunately, the expected size of $\tilde{V}$ sampled according to  \eqref{eqnIdeal} is guaranteed to be sub-linear:
\begin{equation} \label{eqnSize}
\E[|\tilde{V}|] = \sum_{v \in V}g(\{v\}) = \E\left[\sum_{v \in V}f(\{v\},U)\right] \le \E\left[\sum_{v \in V}p(d(v,U))\right] \le n^\beta, 	
\end{equation}
where both inequalities relied on Assumption \ref{assuSublinear}.

\section{Our Approach in Detail: Locality-Sensitive Sampling} \label{secLSS}

To recap, the main conclusion drawn from the previous section is that the ability to sample according to \eqref{eqnIdeal} in sub-linear time is sufficient for our goal of constructing an algorithm that is itself sub-linear and that achieves a constant approximation of $\OPT$. In this section, we will first describe our solution to this sampling problem, which makes use of the abstract $\ANN$ data structures assumed to exist in Assumption \ref{assuANN}. We will then, as a slight detour (which may be skipped), describe what a concrete version of this approach looks like using actual locality-sensitive hash functions. Finally, we close the loop and provide our theoretical guarantee in the form of Theorem \ref{thmGuarantee}. 

\subsection{Approximating the Ideal Sampling Distribution via Locality-Sensitive Sampling}
Now with the goal of executing the ideal sampling distribution \eqref{eqnIdeal}, the brute-force method to sample exactly from this distribution would be to generate $n$ independent Bernoulli variables, whose means are $g(\{v\})$ for each $v \in V$. By Lemma \ref{lemPrune}, repeating this procedure $s = \Omega(k \log k)$ times yields a pruned set of items that preserves a good approximation of $\OPT$ in expectation. However, 
generating the $mn$ Bernoulli variables is clearly a linear time procedure.
Fortunately, it is possible in sub-linear time to {\em approximate} \eqref{eqnIdeal}, i.e. the probabilities in \eqref{eqnIdeal} are not matched exactly, but rather just up to a constant: 
\[  \PR(v \in \tilde{V}) \sim g(\{v\}) \;\; \text{for each }  v \in V.\]
The key observation that makes this possible is that while the probabilities of the individual events $\{v_j \in \tilde{V}\}$ need to be strictly controlled, these events are allowed to be arbitrarily correlated. This is precisely what allows us to leverage the underlying metric space, along with the approximate near neighbor data structures assumed by Assumption \ref{assuANN}, to approximately perform \eqref{eqnIdeal}.

To see why this is possible, first note that allowing arbitrary correlations implies that to sample each $v$ with probability $g(\{v\}) = \E[f(\{v\},U)]$, it suffices to first draw $u$ according to $U$, and then sample each $v$ with probability $f(\{v\},u)$. Next recall that by Assumption 3, there exists a function $p(\cdot)$ such that $f(\{v\},u) \le p(d(v,u))$. This allows us to use near neighbor queries to do the sampling As a simple example, if $p(x) = \mathds{1}(x \le \gamma)$ for some $\gamma$, then sampling with probability 
$f(\{v\},u)$ is equivalent to returning all $v \in V$ within distance $\gamma$ of $u$, and thus a single approximate near neighbor data structure suffices.

\begin{figure}[h]
\centering
\includegraphics[width=.37\columnwidth]{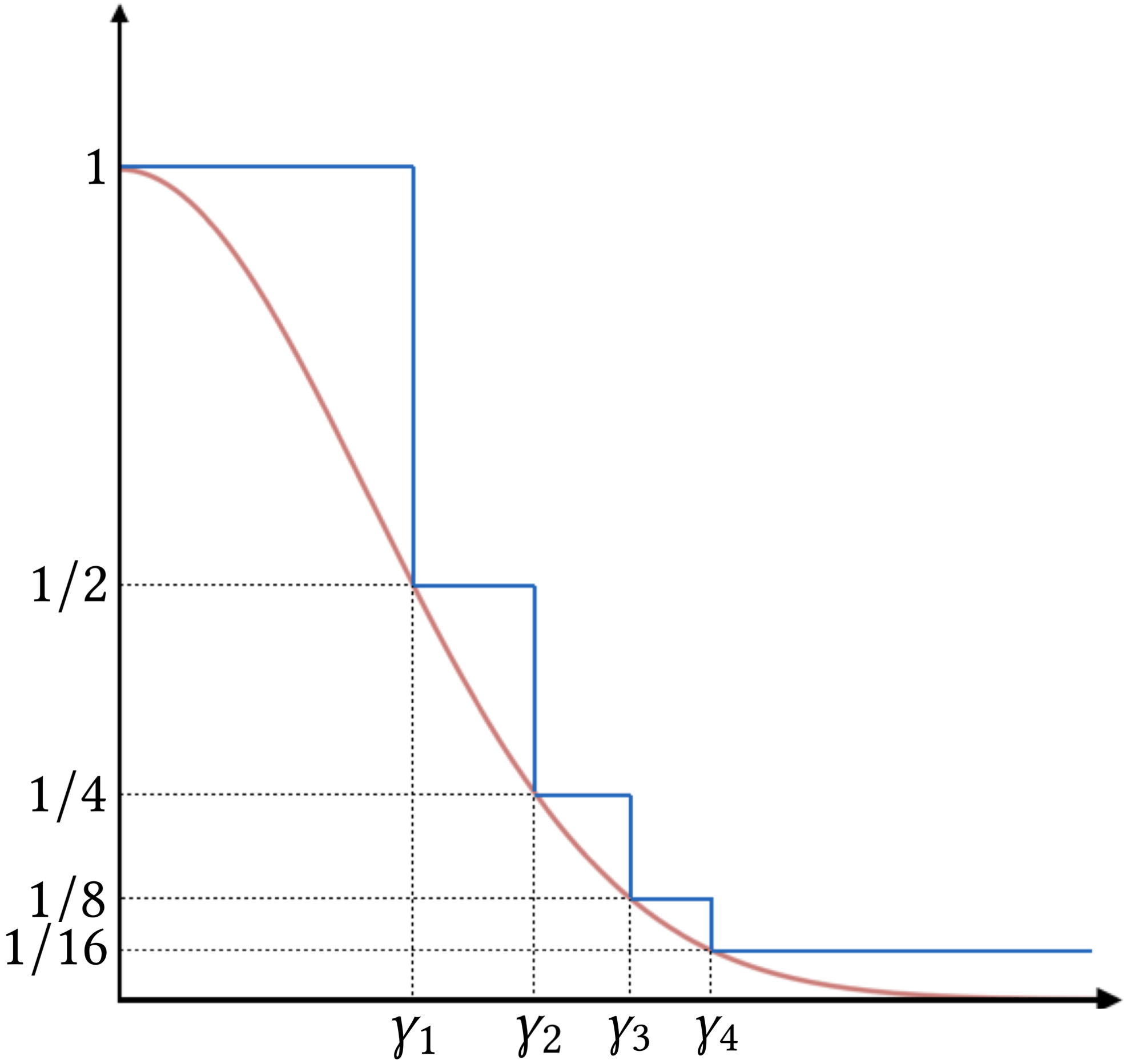}
\caption{Visual depiction of the approximate sampling scheme. The red curve contains the ideal sampling probabilities $p(\cdot)$, and the blue curve shows how we attempt to approximate it using a step function. In this example, $R=4$.}
\label{figDiagram}
\end{figure}

More generally, 
our algorithm utilizes $R =  \lfloor \log_{2} n^{1-\beta} \rfloor$ of these structures to approximate any non-increasing function $p(\cdot)$. For each $r \in [R]$, let 
\begin{equation} \label{eqnRhoGamma}
	\rho_r = \begin{cases} 1/(2^{r}-1), \;\; r \in [R-1]\\ 1/2^{r-1},\;\;\;\;\;\;\; r=R \end{cases}\;\; \text{ and } \;\; \gamma_r = \sup \{x : p(x) \ge 1/2^r).
\end{equation}
See Figure \ref{figDiagram} for a visual depiction of these parameters and our strategy, which is to approximate $p(\cdot)$ by a step function, each step represented by a single near neighbor data structure. 

Our overall scheme then, defined formally below, is to create a set of approximate near neighbor structures, and sample from $p(d(v,u))$ by querying each structure and returning their union. The various parameters for these structures are given by the $\rho_r$'s and $\gamma_r$'s, along with our choice of $\epsilon = 1/2$ (chosen arbitrarily to save on notation). 

\begin{definition}[Locality-Sensitive Sampling]
Let $V$ be a finite subset of a 	metric space $\Mscr$ satisfying Assumption \ref{assuANN},  For any non-increasing function $p:\R^+ \to [0,1]$, and any constant $c > 1$, the \emph{Locality-Sensitive Sampling} data structure is a (randomized) map $\LSS[V,p,c,\beta]: \Mscr \to 2^V$:
\[ \LSS[V,p,c,\beta](u) =  \rho_0V  \ \bigcup  \ \left( \bigcup_{r=1}^R \ANN[\rho_rV,\gamma_r,c,\beta,1/2](u) \right) \;\; \text{ for all } u \in \Mscr, \]
where the $\rho_r$ and $\gamma_r$ are defined as in \eqref{eqnRhoGamma}, $\rho_0 = \frac{1}{2}n^{\beta -1}$, and each $\rho_r V$ denotes a random subset of $V$ gotten by including each element of $V$ independently with probability $\rho_r$.
\end{definition}

The locality-sensitive sampling data structure achieves the approximate sampling distribution we seek in sub-linear time. This is stated formally in the following Lemma, whose proof appears in Appendix \ref{secProofLemLSS}.

\begin{lemma}\label{lemLSS}
For each $u \in \Mscr$ and $v \in V$,
\[ \PR(v \in \LSS[V,p,c,\beta](u)) \ge p(d(v,u))/2.\]
Moreover, each query $\LSS[V,p,c,\beta](u)$ has runtime 
\[ O\left( n^\alpha \log n \right),  \]
where $\alpha = \alpha(c,\beta,1/2)$.
\end{lemma}

\subsection{Aside: LSS Using Locality-Sensitive Hash Functions}
So far, we have assumed the existence of ANN data structures satisfying Assumption \ref{assuANN}, without describing how any of these work. In this subsection, we describe one existing approach based on locality-sensitive hash (LSH) functions (originally described in the seminal work of \cite{indyk1998approximate}), and show how an $\LSS$ structure can be constructed from scratch from these functions. This subsection can be safely skipped without loss of continuity.

The key component of LSH algorithms are {\em LSH families}: let $\Hscr$ be a family of functions defined on $\Mscr$ such that when $h$ is chosen uniformly at random from $\Hscr$, we have $\PR(h(u_1) = h(u_2)) = q(d(u_1,u_2))$. Here, $q: [0,\infty) \to [0,1]$ is some non-increasing function such that $q(0) = 1$ and $q(x) > 0$ if $p(x) > 0$. We will show that sampling from $V$ can be approximated in sublinear time using an LSH family $\Hscr$ and our Locality-Sensitive Sampling procedure: 

\begin{proposition} \label{propLSS}
Let $\Hscr$ and $q$ be defined as in the preceding text, and suppose that 
\[\log_{q(cx)} q(x) \le \delta \;\; \text{ for all } x \text{ and some } \delta < 1.\footnote{We follow the convention that $\log_0 x = 0$ for any $x \in [0,1]$.}\]
Then there exists a locality-sensitive sampling data structure built from these hash functions such that 
Lemma \ref{lemLSS} holds with \[\alpha = \beta + \delta(1-\beta).\]
\end{proposition}
The proof can be found in Appendix \ref{secProofPropLSS}. Proposition \ref{propLSS} is only useful assuming the existence of a family of functions $\Hscr$ satisfying the condition in the statement of the Proposition. Does such a family in fact exist? The search and analysis of appropriate families of functions for various metric spaces has been an active area of research. For our own setting, where $\Mscr = \Sf^{d-1}$ and the metric is induced by the $\ell_2$ norm, there are recent results (\cite{terasawa2007spherical,andoni2015optimal,andoni2015practical}) for the {\em cross-polytope} hash family that essentially amounts to randomly rotating a set of pre-defined points on the sphere and hashing each vector to its nearest point. Even simpler is the {\em hyperplane} LSH family where each function corresponds to a single vector, and the function assigns to any vector the sign of its inner product with the defining vector. \cite{charikar2002similarity} show that $\delta$ can be taken to be $1/c$ using this family.

To describe the locality-sensitive sampling procedure, we begin by defining the vanilla LSH data structure, which we parameterize by $\rho \in [0,1]$ and integers $a,b > 0$. To construct the data structure, first a random subset $\rho V \subset V$ is taken by including each element of $V$ independently with probability $\rho$. Then a total of $b$ hash tables are constructed, with each table storing all of the items in $\rho V$. The hash function for each table $j=1,\ldots,b$ is vector-valued, constructed by drawing functions $h^j_1,\ldots,h^j_a$ independently and uniformly at random from $\Hscr$. This entire construction is done during the preprocessing phase. Then given a query point $u \in \Mscr$, we hash $u$ in each table and return all collisions: 
\[ \text{LSH}_{\rho,a,b}(u) = \left\{v \in \rho V: (h_1^j(v),\ldots,h_a^j(v)) = (h_1^j(u),\ldots,h_a^j(u)) \;\; \text{for some} \;\; j \in [b] \right\}.  \]
Thus, $\text{LSH}_{\rho,a,b}(u)$ is a random subset of $V$, where the randomness is with respect to the sampling when creating $\rho V$ and selecting the hash functions. 

Our locality-sensitive sampling algorithm then utilizes $R = \lceil 1 +\log_{2} n \rceil$ LSH structures. For each $r \in [R]$, let $\rho_r$ and $\gamma_r$ be defined as in \eqref{eqnRhoGamma}.
Moreover, let 
\[a_r = \left\lceil  \log_{q(c\gamma_r)}2^r n^{\beta-1} \right\rceil \;\; \text{and} \;\; 
b_r = \left\lceil \log(2)2^{-r\delta}n^{\delta(1-\beta)}(1/q(\gamma_r)) \right\rceil . \]
Then given a query $u$, for each $r \in [R]$, we calculate $ \text{LSH}_{\rho_r,a_r,b_r}(u)$ and return their union: 
\[  \rho_0V \bigcup \left( \bigcup_{r=1}^R \text{LSH}_{\rho_r,a_r,b_r}(u) \right) .\] 

\begin{figure}[h!]
\centering
\includegraphics[width=.5\textwidth]{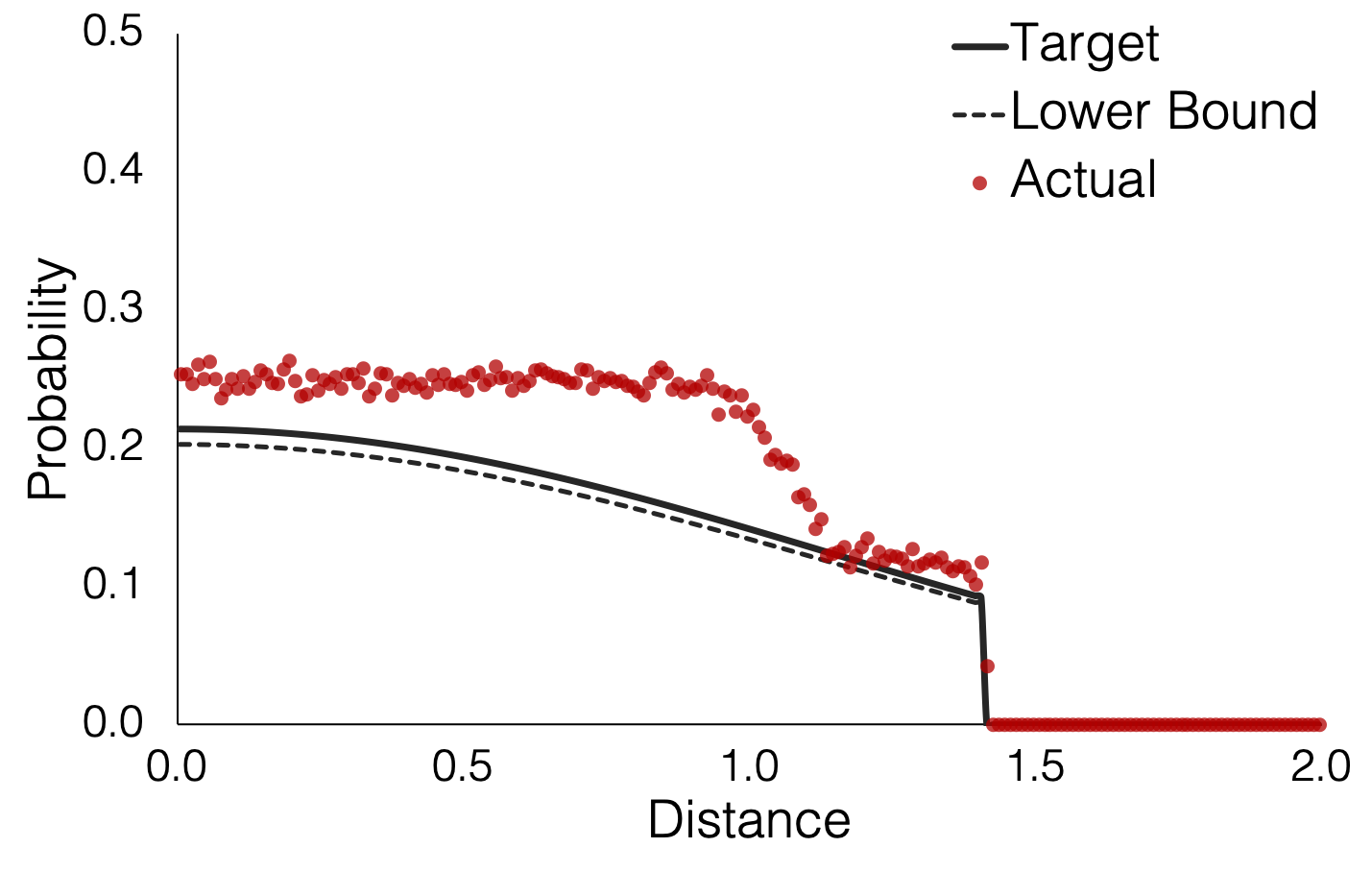}
\caption{Example of sampling of products using Locality-Sensitive Sampling. The sampling probability achieved by the locality sensitive sampling procedure are shown, along with the target sampling distribution and lower bound.}
\label{fig:sample_prob}	
\end{figure}

To demonstrate this procedure concretely, Figure \ref{fig:sample_prob} shows the results of an actual implementation of locality-sensitive sampling on a synthetic dataset (experiments using real data will be described in the following section). This synthetic data consisted of a single query point $u$, and a set $V$ of 50,000 vectors, all lying on the unit Euclidean ball in dimension 50. The vectors in $V$ were randomly generated in such a way that their distance to $u$ is approximately uniform over $[0,2]$.

The target sampling probability function $p(\cdot)$, illustrated by the solid black line, corresponds to the conversion rate for a truncated version of the multinomial logit model.\footnote{The exact choice was \[p(x)= 
\begin{cases}
1 - \frac{10}{10 + \exp(1-x^2/2) } & 0 \leq x< \theta\\
	0 & x \ge \theta \\
\end{cases}\]}
This particular locality-sensitive sampling scheme aims to approximate $p(\cdot)$ by sampling each item at a distance $x$ from $u$ with probability at least $0.95p(x)$, as represented by the dotted line. 
The hash functions used were generated from the aforementioned Hyperplane LSH family.
\footnote{The FALCONN (\cite{andoni2015practical}) software package was used to build the LSH structures} 

To estimate the actual sampling probabilities achieved, we repeated the locality-sensitive sampling procedure 20 times on the dataset (the hash functions were re-chosen randomly in each of these replications leading to different LSH structures), and then measured the fraction of instances for which each item was sampled. 
Each red point in Figure \ref{fig:sample_prob} represents the average sampling probability for a `bin' of about 250 items with nearly equal distance to $u$.
We observe that the locality-sensitive sampling scheme effectively samples as per the desired distribution.

\subsection{Putting It All Together}
Through locality-sensitive sampling, we now have a method of approximating our ideal sampling scheme \eqref{eqnIdeal}. Lemma \ref{lemPrune} requires that $\tilde{V}$ be constructed from $s = \Omega(k \log k)$ {\em independent} samples from this distribution, so we require $s$ instances of this overall structure (each LSS structure itself a combination of ANN structures).

Our final step then is to solve \[ \max_{S \subset \tilde{V}, |S| \le k} g(S) \] using the greedy algorithm, where recall that $g(S) = \E[f(S,U)]$. Specifically, this problem is one of maximizing a monotone submodular set function under cardinality constraint, and as such, is known to admit a $1-e^{-1}$ approximation via a greedy algorithm (\cite{nemhauser1978analysis}). Note that the $1-e^{-1}$ guarantee is the best-known guarantee among polynomial-time algorithms; indeed, even the conversion problem under the no-noise case ($\epsilon=0$) falls into a class of geometric set cover problems known to be APX-hard (\cite{mustafa2014settling}).

The greedy algorithm constructs a solution sequentially as follows: at step $\ell$, having already constructed set $S_{\ell -1}$, we choose $S_{\ell}$ to be
\[S_{\ell} = S_{\ell -1} \cup \argmax_{v \in \tilde{V}} g(S_{\ell -1} \cup \{v\}), \]
where ties are broken arbitrarily. Initiating $S_0$ to be the empty set, the algorithm completes in $k$ steps. Each step of this greedy algorithm requires calculating $g(S_{\ell -1} \cup \{v\})$ for each $v$ in $\tilde{V}$, with each evaluation taking $O(m)$ time. Therefore, the entire greedy procedure runs in $O(km|\tilde{V}|)$ time. To summarize, the last two sections have shown that our algorithm successfully achieves an approximation in sub-linear time.

\begin{theorem} \label{thmGuarantee}
For any $\epsilon_1,\epsilon_2\in(0,1]$, there exists a data structure and algorithm that achieves  
\[(1-e^{-1})[(1-\epsilon_1)\OPT-\epsilon_2]\] 
in expectation and has amortized runtime 
\[O\left((n^\alpha \log n + kmn^\beta)  \frac{k}{\epsilon_2} \log \frac{k}{\epsilon_1}\right),\]
where $\alpha = \alpha(c,\beta,1/2)$.
\end{theorem}

\section{Experiments on Real Data}

We performed two sets of experiments which demonstrate that in real applications:
\begin{enumerate}
\item {\sffamily {\bfseries Modeling:}} our approach to modeling user behavior, particularly with respect to user diversity (i.e. a stochastic user $U$), is more accurate than heuristic sub-linear approaches which do not incorporate diversity.
\item {\sffamily {\bfseries Optimization:}} our algorithm outperforms existing sub-linear heuristics in terms of optimizing reward under our model.
\end{enumerate}

\begin{figure}[h]
\centering
\label{fig:ob}
\includegraphics[width=.4\textwidth]{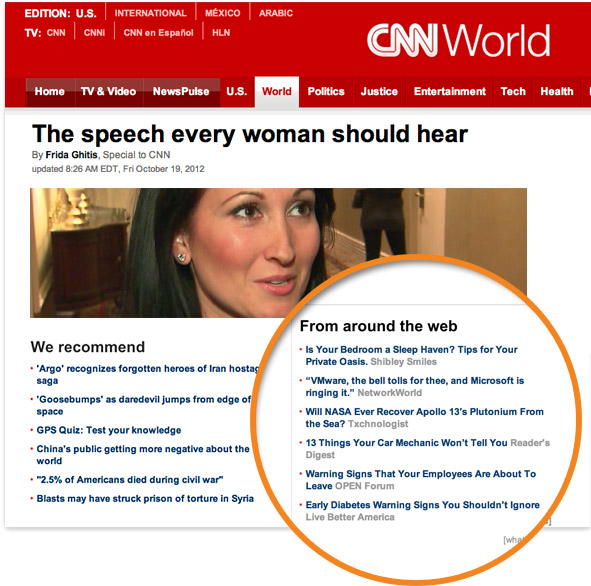}
\caption{Example of content recommendation by Outbrain.}
\label{figOutbrain}
\end{figure}

These experiments were run using real data from Outbrain, an online advertising platform that provides content recommendations on the websites of numerous publishers (e.g. Figure \ref{figOutbrain}).
Outbrain serves over 250 billion personalized content recommendations every month and reaches over 565 million unique visitors. Their promoted articles appear on more than 35,000 websites, reaching over 87\% of internet users in the U.S. (\cite{obw1}). 

Our dataset contains a sample of pages viewed and clicked on by users on multiple publisher sites in the United States over a two-week period. 
Specifically, the data contains about two billion page views for 700 million unique users across 560 websites (amounting to around 100GB of data).
Before both experiments, we performed an initial pruning to remove pages which had been viewed fewer than $750$ times and users who had viewed fewer than $30$ pages. There were approximately $174,000$ pages and $640,000$ users remaining.

We then partitioned the users into two groups, one test group (consisting of 10,000 users) to be used for the actual experiments, and one training group (remaining users) to be used for estimating an accurate metric embedding for the pages. The metric embedding was estimated using a model called \textit{word2vec} (\cite{mikolov2013efficient, mikolov2013distributed}): the pages are treated as `words',  and each user's list of viewed pages (in chronological order) is treated as a `sentence', and the model is trained to predict probabilities of pages appearing near each other (in sentences).\footnote{\textit{word2vec} relies on 
a two layer neural network.  After the training, the weight matrix of the hidden layer of the neural network gives the representation of the words in Euclidean space. 
Its application in this setting is referred to as \textit{prod2vec} (\cite{grbovic2015commerce}). We used the  implementation of  \textit{word2vec} in  Python Machine Learning Library (MLlib), built on Apache Spark.} 
The result is a representation of each page in Euclidean space (which we took to be 50-dimensional), and we normalized each vector to lie on the unit sphere. This embedding was used in both experiments.

Finally, we chose to model user choice as a truncated multinomial logit, meaning  a fixed user type $u$, if recommended a set of pages $S$, has conversion probability
\[ f(S,u) = \frac{\sum_{v \in S, \  v^Tu > 0} \exp(v^\top u / \sigma)}{w+\sum_{v \in S, \ v^Tu > 0}\exp(v^\top u / \sigma)}.	 \]
 Here, $\sigma$ captures the variability of the $\epsilon$'s under the random utility model \eqref{eqnFS} (specifically, it is the scale parameter of the mean-zero Gumbel distribution), and $w$ is a function of the no-choice utility. In both of our experiments, we varied $\sigma$ from 0.01 to 1, and tuned $w$ for realistic conversion probabilities.

\subsection{Modeling Diversity in User Behavior} \label{secExperimentModel}

In the coming second set of experiments, we will consider the recommendation problem assuming that each user $U$ is modeled as the uniform distribution over the first 10 pages he or she has viewed in the data. Before considering that optimization problem though, it is worth asking whether this model for $U$ is reasonable. In particular, is it more accurate than models which fix $U$ to be a single point? (The sub-linear heuristics we will soon compare against can be viewed as implicitly assuming such single point models).

To evaluate the accuracy of any model that is given a user's first 10 pages viewed, we measured how predictive it was of the user's behavior after these first 10 pages viewed.
We compared our mixture model to two single-point models: {\em Mean}, which represents $U$ as the average of the first 10 pages viewed, and {\em Last}, which represents $U$ as the 10th page viewed. Now, for each user, our data contains the pages viewed (the `positive' samples in a classification task), but unfortunately does not specify when pages were offered to the user and not viewed (the `negative' samples).

As a reasonable proxy for a dataset with positive and negative samples, we randomly selected a set of pages assumed to have been offered to each user, in a manner that takes into account the `popularity' of pages.
Specifically, for each page $j$, let $T_j$ denote the number of  views of the web page by the training users. Then, a set of exactly 100 pages was randomly sampled such that the likelihood of each page $j$ being included in the set was proportional to $T_j^\alpha$. The sampling exponent $\alpha$ controls the extent to which higher likelihoods are given to commonly viewed pages. We varied $\alpha$ from 0.2 to 1.0 in our experiments.

\begin{table}[h]
  \centering
    \begin{tabular}{@{}ccccccccccc@{}} \toprule
          &       &       &       & \multicolumn{3}{c}{AUC} &       & \multicolumn{3}{c}{Average Precision} \\  \cmidrule{5-7} \cmidrule{9-11}
    $\sigma$ & $w$   & $\alpha$ &       & Mixed & Mean  & Last    &       & Mixed & Mean  & Last \\ \midrule
    0.01  & 2.75  & 0.2   &       & 0.89  & 0.81  & 0.74  &       & 0.17  & 0.09  & 0.08 \\
          &       & 0.5   &       & 0.89  & 0.81  & 0.73  &       & 0.20  & 0.12  & 0.09 \\
          &       & 0.7   &       & 0.89  & 0.81  & 0.73  &       & 0.25  & 0.14  & 0.11 \\
          &       & 1.0   &       & 0.89  & 0.81  & 0.73  &       & 0.33  & 0.20  & 0.15 \\
          &       &       &       &       &       &       &       &       &       &  \\
    0.1   & 4     & 0.2   &       & 0.95  & 0.95  & 0.90  &       & 0.09  & 0.09  & 0.07 \\
          &       & 0.5   &       & 0.95  & 0.94  & 0.90  &       & 0.11  & 0.11  & 0.09 \\
          &       & 0.7   &       & 0.96  & 0.96  & 0.91  &       & 0.15  & 0.15  & 0.12 \\
          &       & 1.0   &       & 0.96  & 0.95  & 0.91  &       & 0.20  & 0.20  & 0.16 \\
          &       &       &       &       &       &       &       &       &       &  \\
    1     & 500   & 0.2   &       & 0.94  & 0.95  & 0.90  &       & 0.08  & 0.09  & 0.07 \\
          &       & 0.5   &       & 0.94  & 0.95  & 0.90  &       & 0.10  & 0.11  & 0.09 \\
          &       & 0.7   &       & 0.94  & 0.95  & 0.90  &       & 0.13  & 0.14  & 0.12 \\
          &       & 1.0   &       & 0.95  & 0.96  & 0.92  &       & 0.18  & 0.21  & 0.16 \\ \bottomrule
    \end{tabular}
  \caption{Accuracy of our model of user behavior (Mixed) compared against two single-point benchmarks (Mean and Last). The area under the ROC curve (AUC) and average precision are reported for these models, for a variety choices of the multinomial logit tuning parameters ($\sigma,w$) and sampling exponent ($\alpha$). Results are aggregated over the entire set of test users, replicated 20 times each.}
  \label{tabExperimentModel}
\end{table}

Given this set of offered pages for each user, the task for each model was to predict which pages the user had actually viewed. For every page, the models made this prediction by calculating the conversion probability of an assortment containing just that page.
The results are reported in Table \ref{tabExperimentModel}. Since this is effectively a classification task, the metrics reported are the  area under ROC curve (AUC), and the average precision.\footnote{Both are common metrics for classification, lying in $[0,1]$, with higher values signifying greater accuracy.} Table \ref{tabExperimentModel} shows that for small to medium-sized values of $\sigma$, our mixed model more accurately predicts user behavior than the two single-point models, and for large $\sigma$, the accuracy of the mean model is comparable. These results are robust over different choices of $\alpha$. Finally, the actual AUCs of our mixed model run as high as 0.96, which demonstrates that our model is quite accurate in absolute terms.

\subsection{Optimization}
Finally, to test our proposed optimization algorithm, we again modeled users $U$ as being uniformly distributed over their first 10 viewed pages, this time treating these models as ground truth. Over these choice models, we considered the problem of recommending a set of 10 pages to maximize conversion. We compared our own algorithm (LSS) against two benchmarks that are common practice in reality: {\em Mean}, which returns the (approximate) nearest neighbors of the mean of the user's first 10 viewed pages, and {\em Last}, which returns the (approximate) nearest neighbors of the user's 10th viewed page. Both of these benchmarks require a single approximate near neighbor query and are thus sub-linear. 

\begin{table}[h]
  \centering
  
    \begin{tabular}{cccccccccc} \toprule
          &       &       & \multicolumn{3}{c}{Avg. Conversion} &       & \multicolumn{3}{c}{Win Percentage} \\ \cmidrule{4-6} \cmidrule{8-10}
    $\sigma$ & $w$    &       & LSS & Mean  & Last  &       & LSS & Mean  & Last \\ \midrule
    0.01  & 2.75  &       & 0.061 & 0.041 & 0.037 &       & 0.70  & 0.18  & 0.12 \\
          & 2.78  &       & 0.024 & 0.016 & 0.015 &       & 0.67  & 0.19  & 0.15 \\
          &       &       &       &       &       &       &       &       &  \\
    0.1   & 4.0   &       & 0.064 & 0.060 & 0.049 &       & 0.52  & 0.40  & 0.08 \\
          & 4.5   &       & 0.021 & 0.020 & 0.016 &       & 0.52  & 0.38  & 0.09 \\
          &       &       &       &       &       &       &       &       &  \\
    1     & 500   &       & 0.042 & 0.042 & 0.038 &       & 0.25  & 0.74  & 0.01 \\
          & 1000  &       & 0.021 & 0.021 & 0.019 &       & 0.24  & 0.74  & 0.02 \\ \bottomrule
    \end{tabular}%
\caption{Comparison of our algorithm (LSS) to two common practice benchmarks (Mean, Last) on a recommendation problem for mixed multinomial logit users. Each algorithm's conversion rate is reported, averaged over all test users. For each algorithm, the percentage of test users for which that algorithm achieved the highest conversion is also reported.}
  \label{tabExperimentOptimization}%
\end{table}%

The results are summarized in Table \ref{tabExperimentOptimization}. As in the previous set of experiments, we varied $\sigma$ between 0.01 to 1, and tuned $w$ so that the resulting conversion rates were reasonable. Table \ref{tabExperimentOptimization} shows that, for small to medium-sized values of $\sigma$, our algorithm outperforms both benchmarks in terms of both the average conversion rate achieved across all test users, and the proportion of test users for which each algorithm was the best. For large-sized $\sigma$, the mean benchmark is comparable. 

\section{Conclusion}
We proposed a principled approach to offer set optimization that includes (a) a flexible model for user choice that incorporates the underlying structure of commonly estimated item and user metric embeddings, and (b) an algorithm for optimizing offer sets that achieves both a sub-linear runtime and a uniform performance guarantee. Along the way, we developed a sub-linear time algorithm for a certain class of sampling problems that  generalizes the classic approximate near neighbor problem and may be of independent interest. Experiments on a real, large-scale dataset from the online advertising platform Outbrain demonstrated the practicality of our modeling, and superiority of our algorithm against common practice benchmarks.

\bibliographystyle{informs2014}
\bibliography{refs}

\begin{thebibliography}{44}
\providecommand{\natexlab}[1]{#1}
\providecommand{\url}[1]{\texttt{#1}}
\providecommand{\urlprefix}{URL }

\bibitem[{Adomavicius \protect\BIBand{} Tuzhilin(2005)}]{adomavicius2005toward}
Adomavicius G, Tuzhilin A (2005) Toward the next generation of recommender
  systems: A survey of the state-of-the-art and possible extensions. \emph{IEEE
  transactions on knowledge and data engineering} 17(6):734--749.

\bibitem[{Akamai(2017)}]{akamai2017online}
Akamai (2017) Akamai online retail performance report: Milliseconds are
  critical.
  \url{https://www.akamai.com/uk/en/about/news/press/2017-press/akamai-releases-spring-2017-state-of-online-retail-performance-report.jsp}.

\bibitem[{Andoni \protect\BIBand{} Indyk(2008)}]{andoni2008near}
Andoni A, Indyk P (2008) Near-optimal hashing algorithms for approximate
  nearest neighbor in high dimensions. \emph{Communications of the ACM}
  51(1):117.

\bibitem[{Andoni et~al.(2015)Andoni, Indyk, Laarhoven, Razenshteyn,
  \protect\BIBand{} Schmidt}]{andoni2015practical}
Andoni A, Indyk P, Laarhoven T, Razenshteyn I, Schmidt L (2015) Practical and
  optimal lsh for angular distance. \emph{Advances in Neural Information
  Processing Systems}, 1225--1233.

\bibitem[{Andoni \protect\BIBand{} Razenshteyn(2015)}]{andoni2015optimal}
Andoni A, Razenshteyn I (2015) Optimal data-dependent hashing for approximate
  near neighbors. \emph{Proceedings of the Forty-Seventh Annual ACM on
  Symposium on Theory of Computing}, 793--801 (ACM).

\bibitem[{Bentley(1975)}]{bentley1975multidimensional}
Bentley JL (1975) Multidimensional binary search trees used for associative
  searching. \emph{Communications of the ACM} 18(9):509--517.

\bibitem[{Charikar(2002)}]{charikar2002similarity}
Charikar MS (2002) Similarity estimation techniques from rounding algorithms.
  \emph{Proceedings of the thiry-fourth annual ACM symposium on Theory of
  computing}, 380--388 (ACM).

\bibitem[{Ch{\'a}vez et~al.(2001)Ch{\'a}vez, Navarro, Baeza-Yates,
  \protect\BIBand{} Marroqu{\'\i}n}]{chavez2001searching}
Ch{\'a}vez E, Navarro G, Baeza-Yates R, Marroqu{\'\i}n JL (2001) Searching in
  metric spaces. \emph{ACM computing surveys (CSUR)} 33(3):273--321.

\bibitem[{Das et~al.(2007)Das, Datar, Garg, \protect\BIBand{}
  Rajaram}]{das2007google}
Das AS, Datar M, Garg A, Rajaram S (2007) Google news personalization: scalable
  online collaborative filtering. \emph{Proceedings of the 16th international
  conference on World Wide Web}, 271--280.

\bibitem[{Davis et~al.(2013)Davis, Gallego, \protect\BIBand{}
  Topaloglu}]{davis2013assortment}
Davis J, Gallego G, Topaloglu H (2013) Assortment planning under the
  multinomial logit model with totally unimodular constraint structures.
  \emph{Department of IEOR, Columbia University. Available at http://www.
  columbia. edu/~ gmg2/logit\_const. pdf} .

\bibitem[{D{\'e}sir et~al.(2016)D{\'e}sir, Goyal, \protect\BIBand{}
  Segev}]{desir2016assortment}
D{\'e}sir A, Goyal V, Segev D (2016) Assortment optimization under a random
  swap based distribution over permutations model. \emph{EC}, 341--342.

\bibitem[{D{\'e}sir et~al.(2015)D{\'e}sir, Goyal, Segev, \protect\BIBand{}
  Ye}]{desir2015capacity}
D{\'e}sir A, Goyal V, Segev D, Ye C (2015) Capacity constrained assortment
  optimization under the markov chain based choice model. \emph{Operations
  Research, Forthcoming} .

\bibitem[{Farias et~al.(2013)Farias, Jagabathula, \protect\BIBand{}
  Shah}]{farias2013nonparametric}
Farias VF, Jagabathula S, Shah D (2013) A nonparametric approach to modeling
  choice with limited data. \emph{Management Science} 59(2):305--322.

\bibitem[{Freund et~al.(2003)Freund, Iyer, Schapire, \protect\BIBand{}
  Singer}]{freund2003efficient}
Freund Y, Iyer R, Schapire RE, Singer Y (2003) An efficient boosting algorithm
  for combining preferences. \emph{Journal of machine learning research}
  4(Nov):933--969.

\bibitem[{Grbovic et~al.(2015)Grbovic, Radosavljevic, Djuric, Bhamidipati,
  Savla, Bhagwan, \protect\BIBand{} Sharp}]{grbovic2015commerce}
Grbovic M, Radosavljevic V, Djuric N, Bhamidipati N, Savla J, Bhagwan V, Sharp
  D (2015) E-commerce in your inbox: Product recommendations at scale.
  \emph{Proceedings of the 21th ACM SIGKDD International Conference on
  Knowledge Discovery and Data Mining}, 1809--1818 (ACM).

\bibitem[{Han et~al.(2019)Han, G{\'o}mez, \protect\BIBand{}
  Prokopyev}]{han2019assortment}
Han S, G{\'o}mez A, Prokopyev OA (2019) Assortment optimization and
  submodularity .

\bibitem[{Indyk \protect\BIBand{} Motwani(1998)}]{indyk1998approximate}
Indyk P, Motwani R (1998) Approximate nearest neighbors: towards removing the
  curse of dimensionality. \emph{Proceedings of the thirtieth annual ACM
  symposium on Theory of computing}, 604--613 (ACM).

\bibitem[{Jin et~al.(2002)Jin, Si, \protect\BIBand{} Zhai}]{jin2002preference}
Jin R, Si L, Zhai C (2002) Preference-based graphic models for collaborative
  filtering. \emph{Proceedings of the Nineteenth conference on Uncertainty in
  Artificial Intelligence}, 329--336 (Morgan Kaufmann Publishers Inc.).

\bibitem[{Jin et~al.(2003)Jin, Si, Zhai, \protect\BIBand{}
  Callan}]{jin2003collaborative}
Jin R, Si L, Zhai C, Callan J (2003) Collaborative filtering with decoupled
  models for preferences and ratings. \emph{Proceedings of the twelfth
  international conference on Information and knowledge management}, 309--316
  (ACM).

\bibitem[{Juan et~al.(2016)Juan, Zhuang, Chin, \protect\BIBand{}
  Lin}]{juan2016field}
Juan Y, Zhuang Y, Chin WS, Lin CJ (2016) Field-aware factorization machines for
  ctr prediction. \emph{Proceedings of the 10th ACM Conference on Recommender
  Systems}, 43--50 (ACM).

\bibitem[{Karatzoglou et~al.(2010)Karatzoglou, Smola, \protect\BIBand{}
  Weimer}]{karatzoglou2010collaborative}
Karatzoglou A, Smola A, Weimer M (2010) Collaborative filtering on a budget.
  \emph{Proceedings of the Thirteenth International Conference on Artificial
  Intelligence and Statistics}, 389--396.

\bibitem[{Kleinberg et~al.(2017)Kleinberg, Mullainathan, \protect\BIBand{}
  Ugander}]{kleinberg2017comparison}
Kleinberg J, Mullainathan S, Ugander J (2017) Comparison-based choices.
  \emph{Proceedings of the 2017 ACM Conference on Economics and Computation},
  127--144.

\bibitem[{K{\"o}k et~al.(2008)K{\"o}k, Fisher, \protect\BIBand{}
  Vaidyanathan}]{kok2008assortment}
K{\"o}k AG, Fisher ML, Vaidyanathan R (2008) Assortment planning: Review of
  literature and industry practice. \emph{Retail supply chain management},
  99--153 (Springer).

\bibitem[{Kunaver \protect\BIBand{} Po{\v{z}}rl(2017)}]{kunaver2017diversity}
Kunaver M, Po{\v{z}}rl T (2017) Diversity in recommender systems--a survey.
  \emph{Knowledge-Based Systems} 123:154--162.

\bibitem[{Liu et~al.(2018)Liu, He, Feng, Nie, Liu, \protect\BIBand{}
  Zhang}]{liu2018discrete}
Liu H, He X, Feng F, Nie L, Liu R, Zhang H (2018) Discrete factorization
  machines for fast feature-based recommendation. \emph{arXiv preprint
  arXiv:1805.02232} .

\bibitem[{Liu et~al.(2014)Liu, He, Deng, \protect\BIBand{}
  Lang}]{liu2014collaborative}
Liu X, He J, Deng C, Lang B (2014) Collaborative hashing. \emph{Proceedings of
  the IEEE conference on computer vision and pattern recognition}, 2139--2146.

\bibitem[{Mikolov et~al.(2013{\natexlab{a}})Mikolov, Chen, Corrado,
  \protect\BIBand{} Dean}]{mikolov2013efficient}
Mikolov T, Chen K, Corrado G, Dean J (2013{\natexlab{a}}) Efficient estimation
  of word representations in vector space. \emph{arXiv preprint
  arXiv:1301.3781} .

\bibitem[{Mikolov et~al.(2013{\natexlab{b}})Mikolov, Sutskever, Chen, Corrado,
  \protect\BIBand{} Dean}]{mikolov2013distributed}
Mikolov T, Sutskever I, Chen K, Corrado GS, Dean J (2013{\natexlab{b}})
  Distributed representations of words and phrases and their compositionality.
  \emph{Advances in neural information processing systems}, 3111--3119.

\bibitem[{Mustafa et~al.(2014)Mustafa, Raman, \protect\BIBand{}
  Ray}]{mustafa2014settling}
Mustafa NH, Raman R, Ray S (2014) Settling the apx-hardness status for
  geometric set cover. \emph{Foundations of Computer Science (FOCS), 2014 IEEE
  55th Annual Symposium on}, 541--550 (IEEE).

\bibitem[{Nemhauser et~al.(1978)Nemhauser, Wolsey, \protect\BIBand{}
  Fisher}]{nemhauser1978analysis}
Nemhauser GL, Wolsey LA, Fisher ML (1978) An analysis of approximations for
  maximizing submodular set functions—i. \emph{Mathematical Programming}
  14(1):265--294.

\bibitem[{Omohundro(1989)}]{omohundro1989five}
Omohundro SM (1989) \emph{Five balltree construction algorithms} (International
  Computer Science Institute Berkeley).

\bibitem[{Outbrain(2017)}]{obw1}
Outbrain (2017) Similar tech.
  \urlprefix\url{https://www.similartech.com/technologies/outbrain}, [Online;
  accessed April 13, 2017].

\bibitem[{Paulev{\'e} et~al.(2010)Paulev{\'e}, J{\'e}gou, \protect\BIBand{}
  Amsaleg}]{pauleve2010locality}
Paulev{\'e} L, J{\'e}gou H, Amsaleg L (2010) Locality sensitive hashing: A
  comparison of hash function types and querying mechanisms. \emph{Pattern
  Recognition Letters} 31(11):1348--1358.

\bibitem[{Rendle(2010)}]{rendle2010factorization}
Rendle S (2010) Factorization machines. \emph{Data Mining (ICDM), 2010 IEEE
  10th International Conference on}, 995--1000 (IEEE).

\bibitem[{Rusmevichientong et~al.(2010)Rusmevichientong, Shen,
  \protect\BIBand{} Shmoys}]{rusmevichientong2010dynamic}
Rusmevichientong P, Shen ZJM, Shmoys DB (2010) Dynamic assortment optimization
  with a multinomial logit choice model and capacity constraint.
  \emph{Operations research} 58(6):1666--1680.

\bibitem[{Schapire \protect\BIBand{} Singer(1998)}]{schapire1998learning}
Schapire WWCRE, Singer Y (1998) Learning to order things. \emph{Advances in
  Neural Information Processing Systems} 10(451):24.

\bibitem[{Sproull(1991)}]{sproull1991refinements}
Sproull RF (1991) Refinements to nearest-neighbor searching in k-dimensional
  trees. \emph{Algorithmica} 6(1):579--589.

\bibitem[{Talluri \protect\BIBand{} Van~Ryzin(2004)}]{talluri2004revenue}
Talluri K, Van~Ryzin G (2004) Revenue management under a general discrete
  choice model of consumer behavior. \emph{Management Science} 50(1):15--33.

\bibitem[{Terasawa \protect\BIBand{} Tanaka(2007)}]{terasawa2007spherical}
Terasawa K, Tanaka Y (2007) Spherical lsh for approximate nearest neighbor
  search on unit hypersphere. \emph{Workshop on Algorithms and Data
  Structures}, 27--38 (Springer).

\bibitem[{Williams(1977)}]{williams1977formation}
Williams HC (1977) On the formation of travel demand models and economic
  evaluation measures of user benefit. \emph{Environment and planning A}
  9(3):285--344.

\bibitem[{Yianilos(1993)}]{yianilos1993data}
Yianilos PN (1993) Data structures and algorithms for nearest neighbor search
  in general metric spaces. \emph{SODA}, volume~93, 311--21.

\bibitem[{Zhang et~al.(2019)Zhang, Yao, Sun, \protect\BIBand{}
  Tay}]{zhang2019deep}
Zhang S, Yao L, Sun A, Tay Y (2019) Deep learning based recommender system: A
  survey and new perspectives. \emph{ACM Computing Surveys (CSUR)} 52(1):1--38.

\bibitem[{Zhang et~al.(2014)Zhang, Wang, Ruan, \protect\BIBand{}
  Si}]{zhang2014preference}
Zhang Z, Wang Q, Ruan L, Si L (2014) Preference preserving hashing for
  efficient recommendation. \emph{Proceedings of the 37th international ACM
  SIGIR conference on Research \& development in information retrieval},
  183--192.

\bibitem[{Zhou \protect\BIBand{} Zha(2012)}]{zhou2012learning}
Zhou K, Zha H (2012) Learning binary codes for collaborative filtering.
  \emph{Proceedings of the 18th ACM SIGKDD international conference on
  Knowledge discovery and data mining}, 498--506.

\end{thebibliography}

\pagebreak
\appendix
\section{Sample Complexity of the Sample Average Approximation} \label{secAppendixSAA}
\begin{lemma} \label{lemSAA}
Let $U_1,\ldots,U_m$ be i.i.d. samples from distribution $U$, and let $S^*_m$ be an optimal solution to
\[ \max_{S \subset V, |S| \le k} \frac{1}{m}\sum_{i\in[m]}f(S,U_i). \]
There exists some universal constant $C$ such that for any $\epsilon \in (0,1]$,
\[\E[f(S^*_m,U)] \ge \OPT - \epsilon \]
as long as 
\[ m \ge C\frac{k \log n}{\epsilon^2}. \]
\end{lemma}

\begin{myproof}[Proof of Lemma \ref{lemSAA}]
This follows from a standard tail bound for bounded (or sub-gaussian, more generally) variables.
Fix any $t > 0$. For any $S \subset V$, the random variable $f(S,U)$ lies in the interval $[0,1]$ by assumption, and so by Hoeffding's inequality,
\[ \PR\left( \left| \frac{1}{m}\sum_{i\in[m]}f(S,U_i) - \E[f(S,U)] \right| > t  \right) \le 2 \exp\left(- cmt^2 \right),   \]
for some universal constant $c$. Applying a union bound over all cardinality-constrained $S \subset V$, we obtain:
\begin{align*}
\PR\left(\max_{S \subset V, |S| \le k} \left| \frac{1}{m}\sum_{i\in[m]}f(S,U_i) - \E[f(S,U)] \right| > t  \right) 
& \le \sum_{S \subset V, |S| \le k} \PR\left( \left| \frac{1}{m}\sum_{i\in[m]}f(S,U_i) - \E[f(S,U)] \right| > t  \right) \\
& \le 2kn^k \exp\left(- cmt^2 \right), \\
\end{align*}
which implies (e.g. by integrating over $t \ge 0$) that for some constant $C$,
\[ \E\left[ \max_{S \subset V, |S| \le k} \left| \frac{1}{m}\sum_{i\in[m]}f(S,U_i) - \E[f(S,U)] \right| \right] \le C \sqrt{\frac{k\log n}{m}}. \]
The error incurred by optimizing over the sample average approximation is at most twice this uniform bound, which equals $\epsilon$ for $m$ as given in the statement of the theorem. 
\end{myproof}

\section{Additional Proofs}

\subsection{Proof of Lemma \ref{lemLSS}} \label{secProofLemLSS}
First, fix any $u \in \Mscr$ and $v \in V$, and let $r_0 = \lceil-\log_2 p(d(v,u))\rceil$. 
If $p(d(v,u)) \leq 2\rho_0 $, then clearly $$\PR(v \in \LSS[V,p,c](u))  \ge \ \rho_0  \geq \ p(d(v,u))/2.$$
Now, let us consider the case when $p(d(v,u)) > 2\rho_0 $.
Here we have
\begin{align*}	
\PR(v \in \LSS[V,p,c](u)) 
& \geq \PR\left( v \in \bigcup_{r=r_0}^R \ANN[\rho_rV,\gamma_r,c,1/2](u) \right)\\
&= 1- \prod_{r=r_0}^R \PR\left( v \notin \ANN[\rho_rV,\gamma_r,c,1/2](u) \right)\\
&\ge 1- \prod_{r=r_0}^R (1-\rho_r/2) \\
&\ge \frac{1/2}{2^{r-1}} \\
&\ge p(d(v,u))/2,
\end{align*}	
which is precisely the first statement. 

Now, by Assumption \ref{assuANN}, we can guarantee $O(Rn^\alpha)$ runtime as long as the expected number of items returned is $O(n^\beta)$. This is indeed the case:
\begin{align*} 
\sum_{r=1}^R \rho_r \sum_{v\in V}\mathds{1}\{d(v_j,u) \le c \gamma_r \} 
&= \sum_{v \in V} \sum_{r=1}^R \rho_r \mathds{1}\{d(v_j,u) \le c \gamma_r \} \\
&\le \sum_{v \in V}  \max\left\{2p\left(\frac{d(v,u)}{c}\right),\frac{1}{n}\right\} \\
&\le 2n^\beta	
\end{align*}
\qed

\subsection{Proof of Proposition \ref{propLSS}} \label{secProofPropLSS}
Two facts follow from the choice of parameters in the LSH data structures. First, for any $v$ such that $p(u,v) \in (\rho_r/2,\rho_r]$, the probability that the algorithm returns $v$ is at most $\rho_r$ and at least  $\rho_r[1-(1-q(\gamma_r)^{a_r})^{b_r}].$ Since we have
\begin{align*}
 (1-q(\gamma_r)^{a_r})^{b_r} 
 & \le \exp\left(-q(\gamma_r)^{a_r} b_r\right) \\
 & \le \exp\left(- q(\gamma_r) (2^r n^{\beta-1})^{\log_{q(c\gamma_r)} q(\gamma_r)}  \log(2)2^{-r\delta}n^{\delta(1-\beta)}(1/q(\gamma_r))\right) \\
 & \le 1/2, 	
\end{align*}
it follows that this satisfies the sampling requirement. 

Second, the expected total number of collisions in a given LSH structure is at most 
\begin{align*}
(2n^\beta + \rho_r nq(c\gamma_r)^{a_r})b_r 
&\le (2n^\beta + \rho_rn2^rn^{\beta-1} )b_r \\
& = 4n^{\beta}b_r \\
&\le 2^{2-r\delta}n^{\beta+\delta(1-\beta)} (1/q(\gamma_r))\log(2) + 4n^\beta.
\end{align*}
The first inequality follows from the definition of $a_r$, which implies that
\[a_r \ge \log_{q(c\gamma_r)}2^r n^{\beta-1}. \]
The equality comes from the definition of $\rho_r$ and combining terms.
The second inequality follows from the definition of $b_r$, which implies that 
\[b_r \le  \log(2)2^{-r\delta}n^{\delta(1-\beta)}(1/q(\gamma_r)) +1. \]

Thus, across all structure the expected number of collisions is $O(n^{\beta + \delta(1-\beta)} \log n).$

\end{document}